\documentclass[letterpaper, 10 pt, conference]{ieeeconf}  

\let\labelindent\relax

\IEEEoverridecommandlockouts                              

\usepackage{graphics} 
\usepackage{epsfig} 
\usepackage{mathptmx} 
\usepackage{times} 
\usepackage{amsmath} 
\usepackage{amssymb}  
\usepackage{amstext} 
\usepackage{dsfont}

\title{\LARGE \bf
Watch Less, Feel More: Sim-to-Real RL for Generalizable Articulated Object Manipulation via Motion Adaptation and Impedance Control
}

\author{Tan-Dzung Do$^{1,2}$, Nandiraju Gireesh$^{1,2}$, Jilong Wang$^2$, and He Wang$^{1,2,\dagger}$
\thanks{$^{\dagger}$corresponding to \href{mailto:hewang@pku.edu.cn}{hewang@pku.edu.cn}}%
\thanks{$^{1}$CFCS, School of Computer Science, Peking University \quad
$^{2}$Galbot}%
}

\usepackage{mathtools}
\usepackage{amsmath}

\usepackage{amssymb}
\usepackage{amsfonts}
\usepackage{amsopn}
\usepackage{graphicx} 
\usepackage{textcomp}
\usepackage{xfrac}
\usepackage{bbm}
\usepackage{overpic}
\usepackage{subfig}
\usepackage{wrapfig}
\usepackage[ruled,vlined,linesnumbered]{algorithm2e}
\SetAlFnt{\small}
\SetAlCapFnt{\small}
\SetAlCapNameFnt{\small}
\SetAlCapHSkip{0pt}

\usepackage{multirow}

\usepackage{booktabs}
\usepackage{footnote}
\usepackage{paralist}
\usepackage{enumitem}
\usepackage{autobreak}
\usepackage{hyperref}
\usepackage{cleveref}
\usepackage{cite}
\usepackage{siunitx}

\usepackage{color}
\definecolor{green}{rgb}{0, 0.4, 0}
\definecolor{orange}{rgb}{0.8, 0.6, 0.2}
\definecolor{red}{rgb}{1.0, 0.0, 0.0}
\definecolor{teal}{rgb}{0.0, 0.4, 0.4}
\definecolor{purple}{rgb}{0.65,0,0.65}
\definecolor{saffron}{rgb}{0.95,0.75,0.2}
\definecolor{turquoise}{rgb}{0.0,0.5,0.5}
\definecolor{brown}{rgb}{0.5, 0.16, 0.16}

\usepackage{overpic}
\usepackage{currfile} 

\newlength\savedwidth

\definecolor{lightgray}{rgb}{0.6, 0.6, 0.6}

\newcommand{\addcite}[1]{{\textcolor{red}{[cite]}}}

\definecolor{revisedcolor}{RGB}{100,0,200}

\usepackage[normalem]{ulem}

\newcommand{\hidecomment}[1]{}

\usepackage{xspace}

\usepackage[table]{xcolor}
\begin{document}

\let\oldtwocolumn\twocolumn
\renewcommand\twocolumn[1][]{%
    \oldtwocolumn[{#1}{
    \begin{flushleft}
   \centering
    \vspace{0.2cm}
    \vspace{-5mm}
    \includegraphics[width=\textwidth]{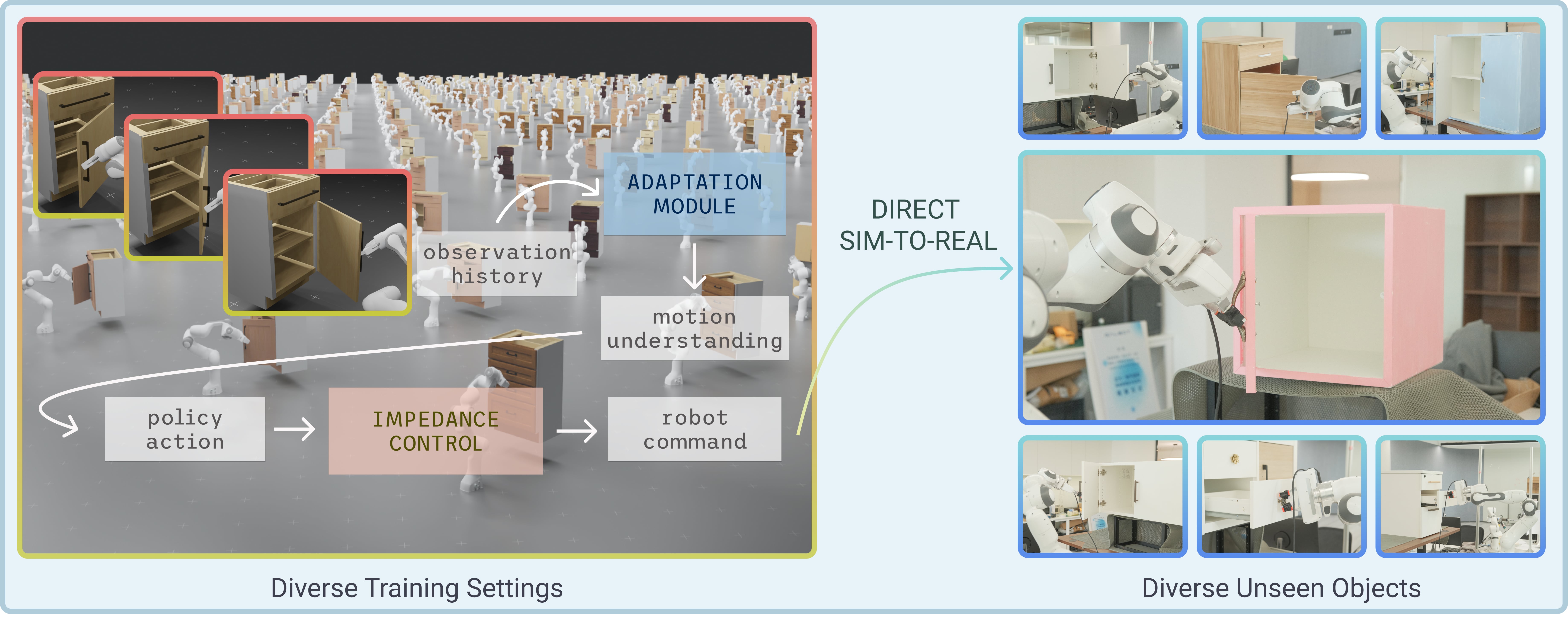}
    \vspace{0.2cm}

    \captionof{figure}{
    We train an RL policy to open doors and drawers in simulation that adapts its action according to the motion of objects by leveraging history observations (left). We directly transfer this policy to reach 80\% joint limit in the real world with closed-loop variable impedance control and achieve 84\% success rate, using only one first-frame RGBD image (right). 
    }\label{fig:teaser}
    \end{flushleft}
    }]
}

\maketitle
\thispagestyle{empty}
\pagestyle{empty}


\begin{abstract}

Articulated object manipulation poses a unique challenge compared to rigid object manipulation as the object itself represents a dynamic environment. In this work, we present a novel RL-based pipeline equipped with variable impedance control and motion adaptation leveraging observation history for generalizable articulated object manipulation, focusing on smooth and dexterous motion during zero-shot sim-to-real transfer (Fig. \ref{fig:teaser}). To mitigate the sim-to-real gap, our pipeline diminishes reliance on vision by not leveraging the vision data feature (RGBD/pointcloud) directly as policy input but rather extracting useful low-dimensional data first via off-the-shelf modules. Additionally, we experience less sim-to-real gap by inferring object motion and its intrinsic properties via observation history as well as utilizing impedance control both in the simulation and in the real world. Furthermore, we develop a well-designed training setting with great randomization and a specialized reward system (task-aware and motion-aware) that enables multi-staged, end-to-end manipulation without heuristic motion planning. To the best of our knowledge, our policy is the first to report 84\% success rate in the real world via extensive experiments with various unseen objects. Webpage: \href{https://watch-less-feel-more.github.io/}{https://watch-less-feel-more.github.io/}

\end{abstract}

\section{INTRODUCTION}

A generalist robot represents a big milestone for the robot learning community, with the potential to revolutionize our daily life. With the ubiquity of articulated objects in both household and industry settings, learning how to efficiently manipulate them is one of the main challenges to achieving this goal. Amid the great progress in the embodied AI field in these couple of years\cite{zhang2023gamma, ha2024umionlegs, forcecontrolepfl, agilebutsafe}, generalizable articulated object manipulation remains an open question due to various reasons. One major challenge is that the true articulation characteristics (e.g. pivot center, friction, stiffness) could only be identified after physical contact is made. For instance, two objects might appear identical but their physical properties differ significantly. As a result, in order to achieve a generalizable articulated object manipulation pipeline that can seamlessly interact with unseen objects, it necessitates a closed-loop pipeline that can adaptively infer these characteristics during the manipulation stage. Another difficulty lies in the joint constraints of objects which require the applied actions to comply with the actual object joint motion. If the robot actions do not tolerate object joint motion and prioritize completing the given commands, it could result in large forces and damage to both objects and the robot.


Recent articulated object manipulation works often rely on visual information as the dominant input for their pipelines. Some prior works leverage vision input in the first frame, either in the form of pointcloud~\cite{coarse,gapartnet,eisner2022flowbot3d} or RGB images~\cite{rgbmanip,dooropen,vrb,where2act, where2explore, gapartnet}, to predict actionable parts followed by a sequence of actions or a waypoint trajectory. This sequence or waypoint is then directly executed in an open-loop manner neglecting all possible physical interaction with objects. This paradigm, despite the natural intuition, overlooks the intrinsic properties of objects and might result in unsafe behaviors. Other works leverage RL backbones \cite{partmanip, umpnet, rlafford, li2024unidoormanip} to output actions in a closed-loop fashion based on vision feedback. However, as this type of pipeline relies heavily on vision feedback at each iteration, it suffers the substantial vision sim-to-real gap inherited from vision modules \cite{dooropen, gapartnet} and can not generalize well. Additionally, during the manipulation stage, this approach might output suboptimal action due to the occlusion of the actionable part. Some \cite{rgbmanip} attempts to leverage impedance control as an off-the-shelf low-level controller to adaptively adjust the predicted waypoint based on some heuristic sample-based rules. However, this approach only affects the local trajectory between two predefined setpoints and results in non-smooth motions.


In this project, we propose combining closed-loop RL with learnable impedance control for generalizable articulated object manipulation. First, we use observation history to manipulate objects in a closed-loop fashion as an alternative for vision input. We evidence our intuition by exemplifying how humans can open a door in the dark: given the information about where the door handle is as well as whether the door is left-hinged or right-hinged, one would estimate the circular motion of the door based on the applied actions and its actual consequential motion. One would then gradually adjust the next actions according to this feedback to complete this task even without direct vision input. We argue that the benefits of leveraging observation history and diminishing reliance on vision, following this intuition, are twofold: 1) By incorporating vision only as a proxy input we can mitigate the vision sim-to-real gap; 2) By leveraging observation and action history, we can implicitly learn the movement of objects, based on the position error after each execution, thus enable a generalizable closed-loop pipeline.

Second, we address the importance of compliant action for articulated object manipulation by introducing variable impedance control to our pipeline. Impedance control is suitable for tasks that require high tolerance to balance setpoint tracking and object joint movement, which fundamentally differentiates articulated object and rigid object manipulation. While implementing a high-frequency variable impedance controller in simulation, we also learn its parameters jointly with our RL policy. We argue that equipping our well-designed training settings with impedance control allows our policy to generate smooth and continuous motions that comply with object joint movements. We find learning motion instead of a single action or discrete waypoints \cite{vatmart, where2act, where2explore} can yield a higher success rate in the real world.

We summarize our contributions as follows:

\begin{itemize}

\item We propose a novel RL-based pipeline for articulated object manipulation with observation and action history as primary inputs while vision only serves as a proxy. (Section \ref{method:distillation}).

\item We design a training setting where each component is realistic for sim-to-real and a reward function system that enables smooth multi-staged end-to-end manipulation without any heuristic motion planning (Section \ref{method:spaces} \& Section \ref{method:reward}).

\item We introduce a variable impedance controller to RL for higher tolerance to object motion, thus benefiting direct sim-to-real transfer (Section \ref{method:impedance}).

\item Through our extensive experiments with 4 tasks and 500 rollouts in the real world, our method's zero-shot inference reaches 96\% and 84\% success rates in simulation and real-world respectively, as well as demonstrates high generalizability to unseen objects.

\end{itemize}


\section{Related work}
\label{sec:relatedworks}
\subsection{Articulated object manipulation}
Manipulating articulated objects is highly challenging due to the wide variety of object geometries and physical properties. Recent works on articulated object manipulation can be broadly categorized into affordance-based and RL-based methods. Affordance-based approaches rely on visual affordance heatmaps~\cite{theory} where each point corresponds to the success rate of manipulation to choose contact points and predict actions~\cite{vatmart,where2act,umpnet,roboabc}.   However, this approach often neglects physical interaction and suffers from large sim-to-real gap \cite{where2act, where2explore, coarse}, which limits their generalizable capability to novel scenes. On the other hand, RL-based methods~\cite{partmanip,rlafford, li2024unidoormanip} with closed-loop feedback have shown better generalization capability. Nevertheless, they utilize point-cloud features as an input to the policy, which makes the exploration space vast and complicates the task. These pipelines also leverage visual input for each inference step which inherently introduces more sim-to-real gap. Our work only leverages low-dimensional vision information captured in the first frame and incorporates history observation during the manipulation stage for better object motion understanding with RL.
 

\subsection{Impedance control for learning-based methods}

Impedance control belongs to the position-force control family where position and force are not decoupled but simultaneously processed, thus enhancing tolerance to feedback force while maintaining a good tracking state. Many contact-rich robotic tasks such as object placement~\cite{industreal} or tool assembly~\cite{admitlearn,atla,factory,fmb,genchip} have successfully demonstrated the compatibility of this type of controller for tasks that consider both position setpoint tracking and object-robot force constraints. For learning-based methods, many works~\cite{manipllm,rgbmanip,imagemanip} introduce impedance control as an off-the-shelf low-level controller for downstream command execution guided by a policy. Some directly incorporate impedance control parameters as learnable variables for RL~\cite{admitlearn,variable}, inverse RL~\cite{impedance_irl}, or analytical optimization methods~\cite{compliancetuning}. These works also showcase that variable impedance control can be more generalizable to different task settings and less labor-expensive than manually tuned impedance control. In this work, we extend the application of impedance control for articulated object manipulation by learning control gain in the simulation and directly transfer to the real world.

\section{Problem Statement}
\label{sec:overview}
\begin{figure*}[ht]
\centering
\begin{overpic}
[width=1.0\linewidth]
{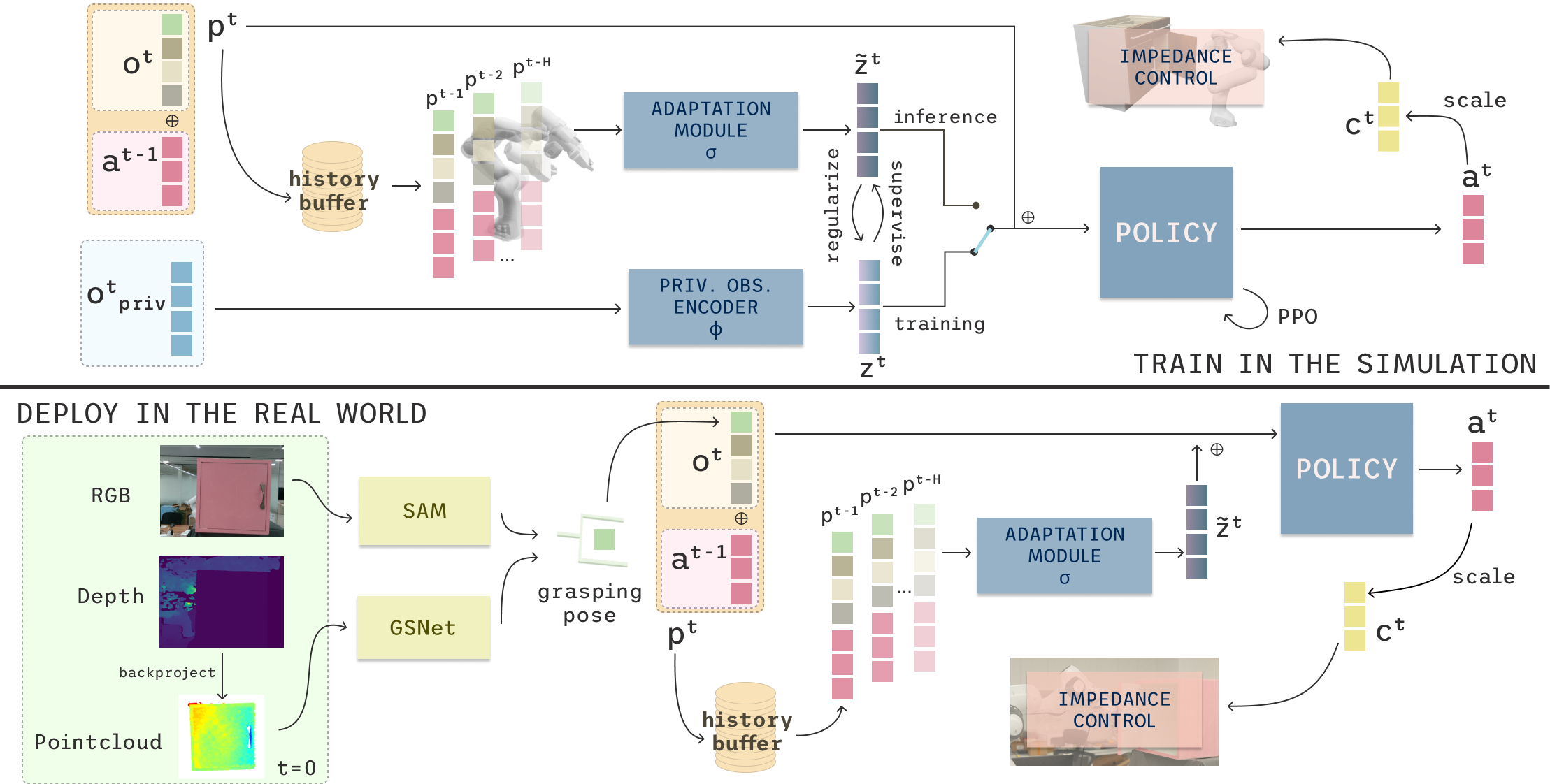}
\end{overpic}
\caption{In the simulation, we train a Privileged Observation Encoder $\phi$ to extract the latent representation of privileged information ${z}^t$ and simultaneously train an Adaptation Module $\sigma$ to infer this representation $\tilde{z}^t$ from $H=10$ previous $(o^t, a^{t-1})$ pairs. The latent representation ${z}^t$ is then concatenated with desired grasping pose $p^t$, robot proprioception $q^t$, robot-object distance $\delta^t$, and categorical object parameters to form policy input. In the real world, we rollout trained policy with Adaptation Module $\sigma$ in an end-to-end manner, executing reaching, grasping, and manipulating. We leverage one RGBD image captured at the first frame to extract the desired grasping pose via off-the-shelf vision modules.}

\vspace{-0.5cm}
\label{fig:pipeline}
\end{figure*}

Given an articulated object \textit{O} and a manipulation task \textit{$\theta$}, we train a policy \textit{$\pi$} to output one dexterous action at a time to finish the task in a closed-loop manner. 

Our task definition is a more challenging and realistic adaptation of VAT-MART\cite{vatmart} and subsequent affordance works  \cite{where2act,where2explore}. Our pulling task (open doors, drawers) requires the policy to reach, grasp actionable parts, and then open untill the object's joint position reaches at least 80\% of the joint limit instead of about half-way\cite{vatmart,rgbmanip}. This criterion, especially when applied to revolute joints, necessitates much dexterous and long-horizon motions since the robot needs to follow the actual $SE(3)$ movements of objects. Moreover, in our settings, we allow only realistic IK configuration of robots (a fixed-base Franka) and do not assume the absolute feasibility of predicted motions as with other waypoint prediction pipelines using a flying gripper or suction cup \cite{where2act,where2explore,umpnet}. 

\section{Proposed method}

\subsection{Action and Observation Space} \label{method:spaces}
We design our framework to facilitate one dexterous action prediction at a time instead of short-horizon primitive actions. Our action for each step $a^t \in \mathbb{R}^{11}$ includes the target delta position $\Delta^t_{xyz} \in \mathbb{R}^3$, target 6D orientation $R^t \in \mathbb{R}^6$, gripper action $G^t \in \mathbb{R}^1$, and impedance control parameter $k_p^t \in \mathbb{R}^1$. Our raw robot action $a^t$ is later converted into robot commands $c^t \in \mathbb{R}^9 $ using an action scaler.

Our observation $o^t$ consists of desired grasping pose $g^t \in \mathbb{R}^7$, robot joint configuration $q^t \in \mathbb{R}^7$, robot-object relative distance $\delta^t \in \mathbb{R}^1$, end-effector pose $ee^t \in \mathbb{R}^9$ with three-dimensional position and 6D rotation, and graspability $ \mathds{1}^t_{grasp} \in \mathbb{R}^1$. Here, desired grasping poses are directly inferred from the handle bounding box in the simulation and from off-the-shelf grasp prediction modules in the real world. Our graspability signal is a distance-based and contact-aware condition, rather than a direct command for open/close gripper. In terms of task-aware observation, for instance, with DoorOpen task, we incorporate noisy pivot center $\tilde{r}^t_{pivot} \in \mathbb{R}^3$, noisy pivot radius $\tilde{r}^t_{radius} \in \mathbb{R}^1$, and right-hinged boolean $\tilde{r}^t_{rh} \in \mathbb{R}^1$. These motion-related arguments serve as high-level guidance for smoother implementation. 
\[ o^t = [g^t, q^t, \delta^t, ee^t, \mathds{1}^t_{grasp}, \tilde{r}^t_{pivot}, \tilde{r}^t_{radius}, \tilde{r}^t_{rh} ] \in \mathbb{R}^{30} \]

Our privileged observation $o^t_{priv}$, including values that are difficult to track in real-world settings, is used only in simulation for better environment understanding. These values are: pivot center ${r}^t_{pivot} \in \mathbb{R}^3$, pivot radius ${r}^t_{radius} \in \mathbb{R}^1$, object stiffness ${r}^t_{stiff} \in \mathbb{R}^1$, object mass ${r}^t_{m} \in \mathbb{R}^1$, object joint position ${q}^t_{obj} \in \mathbb{R}^1$, handle grasped signal $ \mathds{1}^t_{grasped} \in \mathbb{R}^1$.
\[ o^t_{priv} = [{r}^t_{pivot}, {r}^t_{radius},  {r}^t_{m}, {r}^t_{stiff}, {q}^t_{obj}, \mathds{1}^t_{grasped} ] \in \mathbb{R}^{8} \]

\subsection{Online policy distillation with Observation History} \label{method:distillation}
 Articulated object manipulation poses a unique challenge compared to rigid object manipulation because the object itself is a dynamic environment. The fact that object motion can only be observed via physical interactions or that joint ground-truth position is hidden inside the object resembles locomotion tasks where environment parameters (e.g. terrain friction, slope) are difficult to predict. To this end, we adopt the online policy distillation pipeline, which is widely applied for locomotion tasks \cite{deepwholebodycontrol, forcecontrolepfl, agilebutsafe}, and learn two separate modules: Adaptation Module $\sigma$ and Privileged Observation Encoder $\phi$ (Fig. ~\ref{fig:pipeline}).

Privileged Observation Encoder $\phi$ is a shallow MLP, which is utilized during training to learn the latent representation $z^t$ of privileged observations. This 20-dimensional vector is then concatenated with an (observation, action) pair $p^t = (o^t\oplus a^{t-1})$ at the current timestep to form actor inputs. We design the Adaptation Module $\sigma$ to be a temporal architecture to extract latent information about the environment from $H=10$ $p^t$ pairs. We keep only parts of action history as inputs for $\sigma$: position command $\Delta^t_{xyz}$, gripper command $G^t$, and controller gain $k_p^t$.

As the conventional two-staged teacher-student pipeline might result in realizability gap and sim-to-real gap \cite{deepwholebodycontrol}, we simultaneously train Adaptation Module and Privileged Observation Encoder in a single training. Specifically, when jointly train the Adaptation Module with our RL backbone, we also learn to extract similar privileged information $\tilde{z}$ from history buffer by formulating a supervision-regularization loss $\lambda \|z - \text{sg}[\tilde{z}]\|_2 + \|\text{sg}[z] - \tilde{z}\|_2$ on top of PPO objectives ($\text{sg}[.]$ denotes stop gradient operator). We apply a linear schedule for $\lambda$ to prevent our policy from conservative actions in the beginning phase.

\subsection{Reward Design and Domain Randomization} \label{method:reward}
While the proposed framework is adopted widely for locomotion tasks, it remains non-trivial how to transfer this pipeline for fine-manipulation tasks like articulated object manipulation. To facilitate a single end-to-end policy that can efficiently perform multi-staged motions, we introduce stage-conditioned rewards, including task-aware rewards and motion-aware rewards (see Table \ref{table:reward}). 

Task-aware rewards focus on executing a proper motion sequence, complying A-then-B order, rather than cheating to gain success rewards immediately. For instance, at timestep $t$, state $s_1^t$ with the door opened and the door handle grasped firmly by the gripper is rewarded significantly more than state $s_2^t$ without the grasped handle. 

Motion-aware rewards encourage our policy to generate smooth motions while maintaining a high success rate. These terms are often activated after the policy is trained to complete the main task, thus acting as a fine-tuning incentive for smoother execution. We argue that incorporating these regularization terms is crucial and helps bridge the sim-to-real gap by preventing unnecessary motion or non-achievable target poses.

Recent manipulation works \cite{partmanip,gapartnet,rgbmanip} demonstrate that training a policy with domain randomization may benefit sim-to-real transfer. In our work, we mainly focus on tackling the physics gap by asking our policy to understand object motion by object-robot interaction with noisy intrinsic properties. We randomize object positions and object yaw rotations during training to cover a reasonable workspace for real-world settings. In terms of physical intrinsic, we vary the joint friction, stiffness, and mass for more robust sim-to-real transfer. For desired grasping poses, after we infer a pose from part bounding boxes, we introduce random noise along $y$ and $z$ axes, together with a random rotation target from a pre-defined spherical cone.
\begin{table}[]
\centering
\begin{tabular}{lll}
\hline
\multicolumn{1}{c}{\textbf{Term}} & \multicolumn{1}{c}{\textbf{Formula}} & \multicolumn{1}{c}{\textbf{Weight}} \\ \hline
\multicolumn{3}{c}{Nomenclature}  \\ \hline
$\mathds{1}_{d}$ & \multicolumn{1}{c}{$\delta \leq 0.05$} & \multicolumn{1}{c}{-} \\
$\mathds{1}_{dy}$ & \multicolumn{1}{c}{$0.02 \leq \delta \leq 0.08$} & \multicolumn{1}{c}{-} \\
$\mathds{1}_{g}$ & \multicolumn{1}{c}{$\delta \leq 0.015 \wedge \mathds{1}_{\text{contact}}$} & \multicolumn{1}{c}{-} \\
$\tau$ & \multicolumn{1}{c}{joint torque} & \multicolumn{1}{c}{-} \\
$\dot{q}$ & \multicolumn{1}{c}{joint velocity} & \multicolumn{1}{c}{-} \\
$\text{w}_{len}$ & \multicolumn{1}{c}{episode length weight} & \multicolumn{1}{c}{-} \\
$a_t[y]$ & \multicolumn{1}{c}{action on y axis} & \multicolumn{1}{c}{-} \\
$a_t[z]$ & \multicolumn{1}{c}{action on z axis} & \multicolumn{1}{c}{-} \\
\hline
\multicolumn{3}{c}{Task-aware rewards} \\ \hline
success & \multicolumn{1}{c}{$0.05^{\mathds{1}_{d}} * 0.5^{\mathds{1}_{g}} * \mathds{1}_{s}$} & \multicolumn{1}{c}{40.0} \\
distance & \multicolumn{1}{c}{$\exp({-10*(2\delta^{0.5}}))/2 * 0.8^{\mathds{1}_{g}}$} & \multicolumn{1}{c}{0.6} \\
object state & \multicolumn{1}{c}{$q_{obj} * 0.5^{\mathds{1}_{g}} * 0.5^{\mathds{1}_{d}} * \text{w}_{len}$} & \multicolumn{1}{c}{1.0} \\
grasp & \multicolumn{1}{c}{$0.2 * \mathds{1}_{g}$} & \multicolumn{1}{c}{0.05} \\
\hline
\multicolumn{3}{c}{Motion-aware rewards}  \\
\hline
energy & \multicolumn{1}{c}{$\sum{(\tau \dot{q})^{0.5}} * \mathds{1}_{g}$} & \multicolumn{1}{c}{-0.05} \\
track pos. & \multicolumn{1}{c}{$\exp(-4(c_{pos} - ee_{pos})) * \mathds{1}_{d}$} & \multicolumn{1}{c}{0.025} \\
track rot. & \multicolumn{1}{c}{$\exp(-4\Delta(c_{ori} - ee_{ori})) * \mathds{1}_{d}$} & \multicolumn{1}{c}{0.004} \\
smoothness & \multicolumn{1}{c}{$\sum  \mathds{1}_{[\text{sgn}(a_t) \neq \text{sgn}(a_{t-1})]} * (a_t - a_{t-1})$} & \multicolumn{1}{c}{-0.001} \\
y reg. & \multicolumn{1}{c}{$\mathds{1}_{dy} * (a_t[y]*15)^2$} & \multicolumn{1}{c}{-0.005} \\
z reg. & \multicolumn{1}{c}{$\mathds{1}_{g} * (a_t[z]*15)^2$} & \multicolumn{1}{c}{-0.07} \\
\end{tabular}
\caption{Reward functions}
\label{table:reward}
\end{table}

\subsection{Variable Impedance Control} \label{method:impedance}

The goal of impedance control is to follow a desired trajectory $x_d$ considering the external force $F_{ext}$ resulted from the interaction between the robot and the environment. The design of impedance control follows a mass-spring-damper system that can dynamically adjust target setpoints based on feedback force as well as the stiffness of the environment. The dynamics model of impedance control is:
\[M(\ddot{x_c} - \ddot{x_d}) + D(\dot{x_c} - \dot{x_d}) + K({x_c} - {x_d}) = F_{ext}\]
where $M$ is the mass-inertia matrix of the robot, $D$ is the damping matrix, $K$ is the stiffness matrix, and $[\ddot{x_c}, \dot{x_c}, {x_c}]$ is impedance trajectory outputs.

In our pipeline, we learn to predict the stiffness factor $k_p$ of our Cartesian impedance controller  and expand it into a six-dimensional
diagonal matrix $K$. Following \cite{admitlearn, compliancetuning}, we assume that $M, K, D$ are positive definite diagonal matrices to ensure system stability. To this end, we scale actor prediction $k_p$ by:
\[c^t_{k_p} = \text{clip}(a^t_{k_p}, -1, 1) * 40 + 100\]
We find this value range generates reasonable motions in both simulation and real-world experiments. From stiffness matrix $K$, we then infer the damping matrix with the critical damping condition $D = 2\sqrt{MK}$.

\begin{table*}[ht]
\centering
\setlength{\tabcolsep}{4mm}
\begin{tabular}{c c c c c c c c c c c}
\multicolumn{1}{c}{\multirow{2}{*}{\textbf{Baselines}}} &  \multicolumn{1}{c}{\multirow{2}{*}{\textbf{Type}}} & \multicolumn{2}{c}{\textbf{OpenDoor}} & \multicolumn{2}{c}{\textbf{OpenDrawer}} & \multicolumn{2}{c}{\textbf{OpenDoor+}} & \multicolumn{2}{c}{\textbf{OpenDrawer+}} \\ \cline{3-10}
\textbf{} & \textbf{} & \textbf{Train} & \textbf{Test} & \textbf{Train} & \textbf{Test} & \textbf{Train} & \textbf{Test} & \textbf{Train} & \textbf{Test}\\ \hline
PPO & Closed-loop & 0.04 &  0.05 & 0.09 & 0.11 & 0.02 & 0.02 & 0.03 & 0.02 \\ 
Where2act~\cite{where2act} &  Open-loop & 0.22 & 0.14 & 0.31 & 0.27 & 0.02 & 0.02 & 0.01 & 0.01 \\ 
RGBManip~\cite{rgbmanip} & Closed-loop & 0.62 & 0.59 & 0.63 & 0.67 & 0.38 & 0.41 & 0.49 & 0.42 \\ 
GAPartNet~\cite{gapartnet} & Open-loop & 0.70 & 0.75 & 0.51 & 0.59 & 0.40 & 0.44 & 0.45 & 0.49 \\ 
PartManip~\cite{partmanip} & Closed-loop & 0.75 & 0.70 & 0.83 & 0.77 & 0.68 & 0.57 & 0.62 & 0.59 \\ \hline
\textbf{Ours} & Closed-loop & \textbf{0.96} & \textbf{0.95} & \textbf{0.97} & \textbf{0.96} & \textbf{0.96} & \textbf{0.93} & \textbf{0.97} & \textbf{0.96} \\ 
\end{tabular}
\caption{Comparison with Baselines in Simulation}
\label{table:comparison_baselines}
\end{table*}
\begin{figure*}[ht]
\centering
\begin{overpic}
[width=1.0\linewidth]
{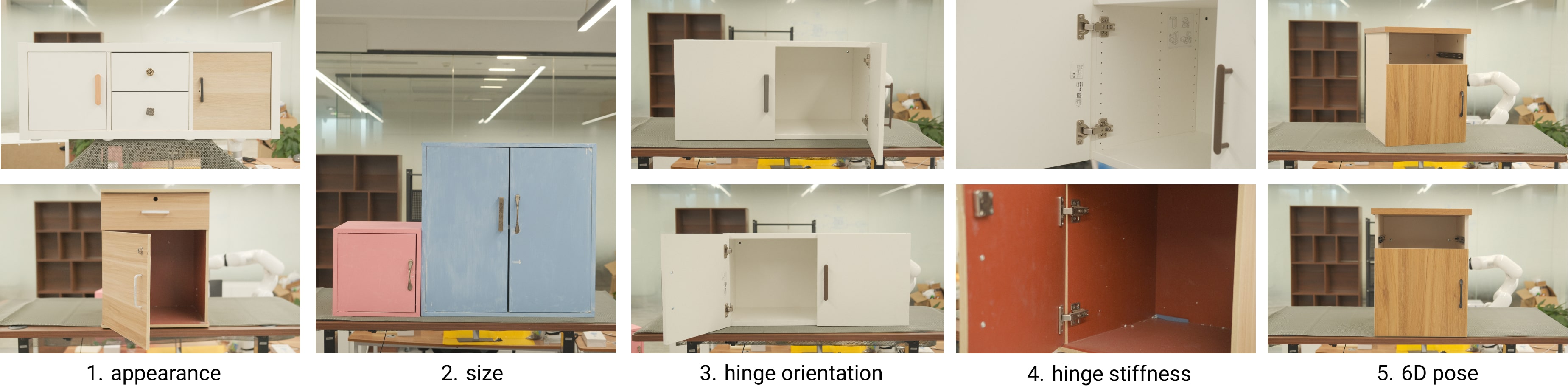}
\end{overpic}
\caption{We extensively evaluate our policy in the real world with a wide range of unseen objects, varied in appearance, size, hinge orientation, and hinge stiffness. We demonstrate our performance in a reasonable workspace, with objects facing front or tilting slightly around the $z$ axis.}

\vspace{-0.5cm}
\label{fig:obj}
\end{figure*}
\section{Experiments}
To verify the effectiveness of the proposed method, we conduct extensive evaluations in both
simulation and real-world settings.

\subsection{Data and Task Settings}
In the simulation, following the settings of PartManip~\cite{partmanip}, we conduct our experiments in the IsaacGym simulator and the large-scale PartNet-Mobility dataset \cite{partnet}. We use a fixed-base Franka and a total of 346 articulated 3D objects covering both doors and drawers (modified \textit{StorageFurniture} subset), to carry out the simulation experiments.

In the real-world setting, we perform experiments with a variety of household objects using the Franka Emika robotic arm equipped with an on-hand RealSense D415 camera to capture RGBD images. We leverage Segment Anything (SAM) \cite{kirillov2023segment} for actionable part pointcloud extraction using a first-framed RGBD image and GSNet \cite{gsnet} for grasp prediction.

We evaluate our proposed pipeline with two following tasks: OpenDoor/OpenDoor+ and OpenDrawer/OpenDrawer+. 

\textbf{OpenDoor/OpenDoor+: }A door is initially closed, the agent needs to open the door larger than 15\%/80\% of the maximum door swing. The key requirement for our task setting is that the gripper should firmly grasp the handle while opening the door without cheating by opening from the side or with the robot body.

\textbf{OpenDrawer/OpenDrawer+: }A drawer is initially closed, the agent needs to open the drawer larger than 20\%/80\% of the maximum opening length. Similar to the \textbf{OpenDoor} task, we require the gripper to firmly grasp the handle while opening the drawer.

For simulation and real-world settings, we adopt Success Rate (SR) as the major evaluation metric.
\begin{table}[]
\centering
\begin{tabular}{c c c c c c c}
\multicolumn{1}{c}{\multirow{2}{*}{\textbf{Methods}}} & \multicolumn{3}{c}{\textbf{OpenDoor+}} & \multicolumn{3}{c}{\textbf{OpenDrawer+}} \\ \cline{2-7} 
                 & \textbf{Train} & \textbf{Test} & \textbf{Real} & \textbf{Train} & \textbf{Test} & \textbf{Real}\\ \hline
W/o Distillation & 0.80 & 0.77 & 0.62 & 0.78 & 0.74 & 0.60 \\ 
W/o Imp. Ctr. & 0.84 & 0.82 & 0.40 & 0.90 & 0.90 & 0.44 \\ 
W/o Reg. & 0.88 & 0.86 & 0.64 & 0.92 & 0.87 & 0.70 \\ 
W/o Rand. & 0.91 & 0.89 & 0.66 & 0.93 & 0.91 & 0.64 \\ \hline
\textbf{Ours} & \textbf{0.96} & \textbf{0.93} & \textbf{0.80} & \textbf{0.97} & \textbf{0.96} & \textbf{0.84} \\ 
\end{tabular}
\caption{Ablation Study and Real-world Performance}
\label{table:ablation}
\end{table}

\begin{figure}[ht]
\centering
\begin{overpic}
[width=0.95\linewidth]
{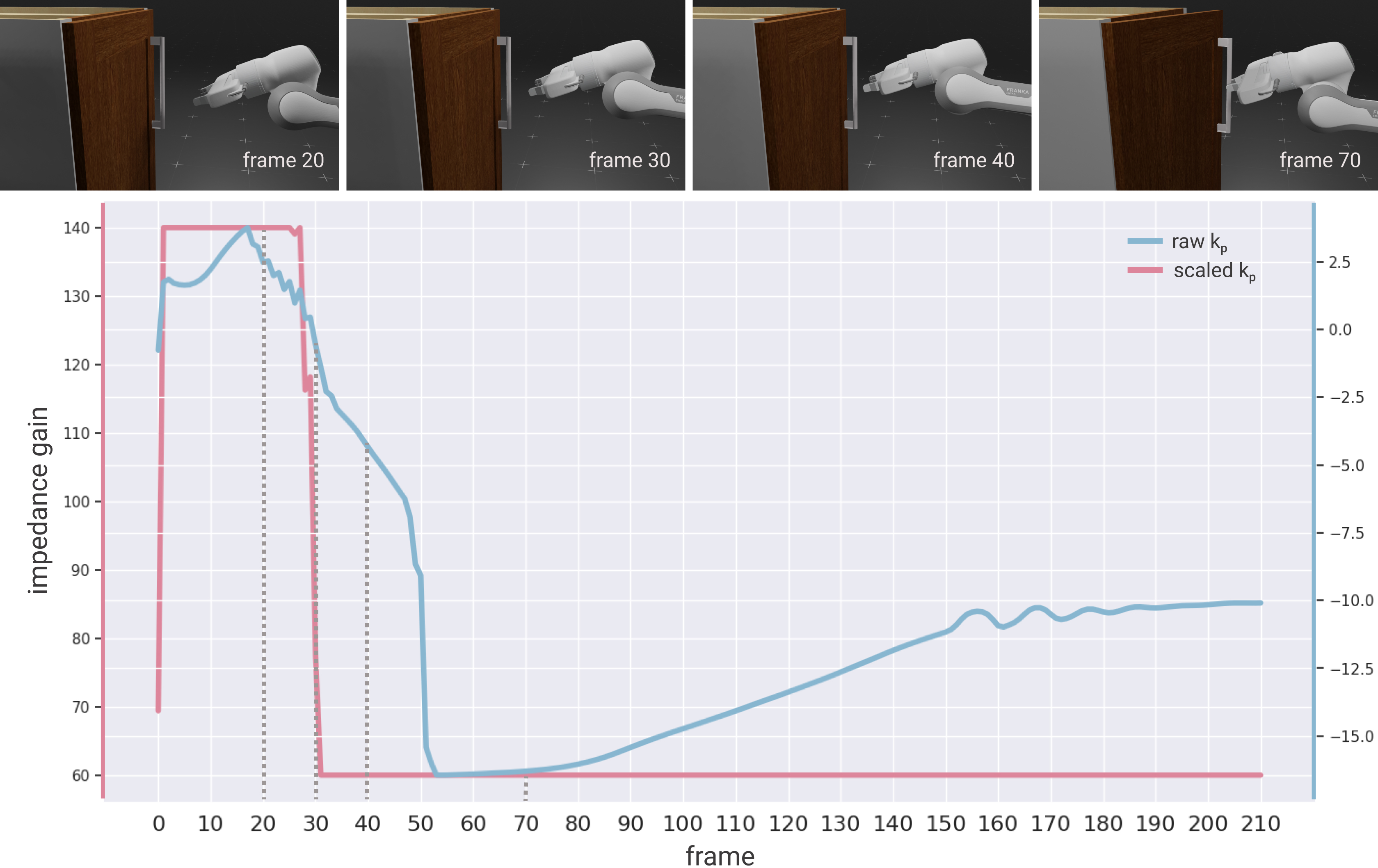}
\end{overpic}
\caption{Our learned controller gain actively adapts to the manipulation stages even without a direct gain reward: stiffer while reaching, softer while opening.}
\vspace{-0.5cm}
\label{fig:kp}
\end{figure}
\subsection{Baselines and Ablation Study Design}
We compare our proposed method with articulated-object manipulation pipelines that follow sim-to-real RL paradigm. 

\textbf{PPO}. We directly use the PPO algorithm to learn a state-based policy to handle each task. The detailed PPO parameters and training strategy are similar to our method.

\textbf{Where2Act~\cite{where2act}}. An affordance learning framework predicting the visual actionable affordance using a partial point cloud. We include the part mask as an additional dimension in our task, while keeping other aspects unchanged. 

\textbf{PartManip~\cite{partmanip}}. A vision-based policy learning method that first trains a state-based expert with part-based canonicalization and part-aware rewards, and then distills the knowledge to a vision-based student policy.

\textbf{RGBManip~\cite{rgbmanip}}. An image-only learning method that leverages an eye-on-hand monocular camera to actively perceive the articulated object from multiple perspectives to enhance 6D pose accuracy.

\textbf{GAPartNet~\cite{gapartnet}}. A vision-based method that first does cross-category part segmentation and pose estimation, and then uses the predicted part poses for heuristic-based manipulation

To highlight the contribution and effectiveness of each module within our approach, we conducted four comprehensive ablation studies:

\textbf{Ours w/o Policy Distillation}. We train a policy with observations from only current timestep $o^t$, omitting Adaptation Module and Privileged Observation Encoder.

\textbf{Ours w/o Variable Impedance Control}. We utilize Cartesian Position Control as low-level controller for our policy.

\textbf{Ours w/o Regularization}. We excluded motion-aware rewards from our reward functions.

\textbf{Ours w/o Randomization}. We exclude all forms of randomization in our pipeline, including object pose, desired grasping pose, friction, stiffness, mass, and noisy intrinsic.

\subsection{Results and Findings}

Results of simulation experiments are shown in Table ~\ref{table:comparison_baselines}, from which we can see that while most baselines perform reasonably well on the training set, their performance tends to decline significantly on the testing set. In contrast, our method maintains consistently strong performance on the evaluation set, without a sharp drop, highlighting the excellent generalization ability of our approach. We also find our controller learns to adapt to different manipulation stages, even without any direct gain rewards (Fig. ~\ref{fig:kp}). Specifically, when the gripper is far from the object, it turns stiffer by setting the controller gain to a higher $k_p$. On the other hand, when the distance is reduced, to minimize the collision penalty, it becomes softer with a lower $k_p$.

Our policy rollout performance in real world can be found in Table ~\ref{table:ablation}. We conduct 50 experiments for our pipeline and each ablated model (500 runs in total) on diverse objects (Fig. ~\ref{fig:obj}). We further investigate our success rate by decoupling the failure cases due to grasp pose estimation in Grasping Stage and due to our pipeline in Opening Stage. For OpenDoor+, we find $6/50$ inferences fail during Grasping Stage while only $4/50$ fail during Opening Stage, suggesting that if a stable grasping pose is initiated, our policy might yield $40/44=0.90\%$ SR. For OpenDrawer+, $7/8$ failure cases are due to unsuccessful grasping.

With the ablation study results demonstrated in Table~\ref{table:ablation}, apart from SR drop in both simulation and the real world, we aim to highlight the non-smooth motions of real-world executions. For \textit{W/o Impedance Control}, we find the main reason for failure cases ($40\%$ drop) is the low flexibility of position control, which requires each predicted action to be executed precisely. This would generate large joint torque to overcome the feedback force of objects, resulting in the robot arm being triggered to stop. In simulation, this behavior does not seem to severely hurt the performance, as evidenced by $>0.8$ success rate. However, in the real world, large torque is substantially dangerous and would trigger an emergency stop, emphasizing the necessity for impedance control. For \textit{W/o Distillation} and \textit{W/o Randomization}, the policy often finishes the task halfway, even when we manually tune a stiffer base value for the impedance controller. We claim that this behavior is due to the physics sim-to-real gap which resulted from non-diverse training settings and short-term observation. For \textit{W/o Regularization}, the reaching and opening motions are jerky, which are highly undesirable and result in grasp failure and contact lost during execution. 

In this work, we hope to introduce a reliable RL policy that can be seamlessly deployed in diverse real-world settings. Our experiments, conducted in both simulation and real-world scenarios, suggest that the manipulation stage should be learned as a smooth and continuous motion in simulation, instead of a discrete waypoint. Together with the tolerance of impedance control, the close-loop real-world transfer could be more efficient, even if the action predictions are slightly suboptimal. 

\section{CONCLUSIONS}
In this work, we introduce a novel RL framework equipped with variable impedance control for end-to-end articulated object manipulation, which adaptively learns the object movement from observation and action history instead of naively executing a trajectory predicted before any robot-object contact. We demonstrate great sim-to-real transfer capability on diverse test objects in the real world and achieve 80\% and 84\% success rate for OpenDoor+ and OpenDrawer+ tasks, respectively, outperforming all existing works. Along with quantitative results, our policy can generate smooth and dexterous motion thanks to our well-designed training settings and reward functions. We hope our work can suggest an alternative way to leverage vision information, as well as other potential modalities (e.g. tactile grasp signal), to better bridge the sim-to-real gap for future RL-based manipulation works.

\section{Acknowledgment}
We thank all reviewers for their insights and suggestions.










\newpage
\bibliographystyle{IEEEtran}
\bibliography{IEEEabrv,reference}

\begin{thebibliography}{10}
\providecommand{\url}[1]{#1}
\csname url@samestyle\endcsname
\providecommand{\newblock}{\relax}
\providecommand{\bibinfo}[2]{#2}
\providecommand{\BIBentrySTDinterwordspacing}{\spaceskip=0pt\relax}
\providecommand{\BIBentryALTinterwordstretchfactor}{4}
\providecommand{\BIBentryALTinterwordspacing}{\spaceskip=\fontdimen2\font plus
\BIBentryALTinterwordstretchfactor\fontdimen3\font minus \fontdimen4\font\relax}
\providecommand{\BIBforeignlanguage}[2]{{%
\expandafter\ifx\csname l@#1\endcsname\relax
\typeout{** WARNING: IEEEtran.bst: No hyphenation pattern has been}%
\typeout{** loaded for the language `#1'. Using the pattern for}%
\typeout{** the default language instead.}%
\else
\language=\csname l@#1\endcsname
\fi
#2}}
\providecommand{\BIBdecl}{\relax}
\BIBdecl

\bibitem{zhang2023gamma}
J.~Zhang, N.~Gireesh, J.~Wang, X.~Fang, C.~Xu, W.~Chen, L.~Dai, and H.~Wang, ``Gamma: Graspability-aware mobile manipulation policy learning based on online grasping pose fusion,'' \emph{arXiv preprint arXiv:2309.15459}, 2023.

\bibitem{ha2024umionlegs}
H.~Ha, Y.~Gao, Z.~Fu, J.~Tan, and S.~Song, ``{UMI} on legs: Making manipulation policies mobile with manipulation-centric whole-body controllers,'' 2024.

\bibitem{forcecontrolepfl}
T.~Portela, G.~B. Margolis, Y.~Ji, and P.~Agrawal, ``Learning force control for legged manipulation,'' \emph{arXiv preprint arXiv:2405.01402}, 2024.

\bibitem{agilebutsafe}
T.~He, C.~Zhang, W.~Xiao, G.~He, C.~Liu, and G.~Shi, ``Agile but safe: Learning collision-free high-speed legged locomotion,'' \emph{arXiv preprint arXiv:2401.17583}, 2024.

\bibitem{coarse}
S.~Ling, Y.~Wang, R.~Wu, S.~Wu, Y.~Zhuang, T.~Xu, Y.~Li, C.~Liu, and H.~Dong, ``Articulated object manipulation with coarse-to-fine affordance for mitigating the effect of point cloud noise,'' in \emph{2024 IEEE International Conference on Robotics and Automation (ICRA)}.\hskip 1em plus 0.5em minus 0.4em\relax IEEE, 2024, pp. 10\,895--10\,901.

\bibitem{gapartnet}
H.~Geng, H.~Xu, C.~Zhao, C.~Xu, L.~Yi, S.~Huang, and H.~Wang, ``Gapartnet: Cross-category domain-generalizable object perception and manipulation via generalizable and actionable parts,'' in \emph{Proceedings of the IEEE/CVF Conference on Computer Vision and Pattern Recognition}, 2023, pp. 7081--7091.

\bibitem{eisner2022flowbot3d}
B.~Eisner, H.~Zhang, and D.~Held, ``Flowbot3d: Learning 3d articulation flow to manipulate articulated objects,'' \emph{arXiv preprint arXiv:2205.04382}, 2022.

\bibitem{rgbmanip}
B.~An, Y.~Geng, K.~Chen, X.~Li, Q.~Dou, and H.~Dong, ``Rgbmanip: Monocular image-based robotic manipulation through active object pose estimation,'' in \emph{2024 IEEE International Conference on Robotics and Automation (ICRA)}.\hskip 1em plus 0.5em minus 0.4em\relax IEEE, 2024, pp. 7748--7755.

\bibitem{dooropen}
H.~Xiong, R.~Mendonca, K.~Shaw, and D.~Pathak, ``Adaptive mobile manipulation for articulated objects in the open world,'' \emph{arXiv preprint arXiv:2401.14403}, 2024.

\bibitem{vrb}
S.~Bahl, R.~Mendonca, L.~Chen, U.~Jain, and D.~Pathak, ``Affordances from human videos as a versatile representation for robotics,'' in \emph{Proceedings of the IEEE/CVF Conference on Computer Vision and Pattern Recognition}, 2023, pp. 13\,778--13\,790.

\bibitem{where2act}
K.~Mo, L.~J. Guibas, M.~Mukadam, A.~Gupta, and S.~Tulsiani, ``Where2act: From pixels to actions for articulated 3d objects,'' in \emph{Proceedings of the IEEE/CVF International Conference on Computer Vision}, 2021, pp. 6813--6823.

\bibitem{where2explore}
C.~Ning, R.~Wu, H.~Lu, K.~Mo, and H.~Dong, ``Where2explore: Few-shot affordance learning for unseen novel categories of articulated objects,'' \emph{Advances in Neural Information Processing Systems}, vol.~36, 2024.

\bibitem{partmanip}
H.~Geng, Z.~Li, Y.~Geng, J.~Chen, H.~Dong, and H.~Wang, ``Partmanip: Learning cross-category generalizable part manipulation policy from point cloud observations,'' in \emph{Proceedings of the IEEE/CVF Conference on Computer Vision and Pattern Recognition}, 2023, pp. 2978--2988.

\bibitem{umpnet}
Z.~Xu, Z.~He, and S.~Song, ``Universal manipulation policy network for articulated objects,'' \emph{IEEE robotics and automation letters}, vol.~7, no.~2, pp. 2447--2454, 2022.

\bibitem{rlafford}
Y.~Geng, B.~An, H.~Geng, Y.~Chen, Y.~Yang, and H.~Dong, ``Rlafford: End-to-end affordance learning for robotic manipulation,'' in \emph{2023 IEEE International Conference on Robotics and Automation (ICRA)}.\hskip 1em plus 0.5em minus 0.4em\relax IEEE, 2023, pp. 5880--5886.

\bibitem{li2024unidoormanip}
Y.~Li, X.~Zhang, R.~Wu, Z.~Zhang, Y.~Geng, H.~Dong, and Z.~He, ``Unidoormanip: Learning universal door manipulation policy over large-scale and diverse door manipulation environments,'' \emph{arXiv preprint arXiv:2403.02604}, 2024.

\bibitem{vatmart}
R.~Wu, Y.~Zhao, K.~Mo, Z.~Guo, Y.~Wang, T.~Wu, Q.~Fan, X.~Chen, L.~Guibas, and H.~Dong, ``Vat-mart: Learning visual action trajectory proposals for manipulating 3d articulated objects,'' \emph{arXiv preprint arXiv:2106.14440}, 2021.

\bibitem{theory}
J.~Gibson, ``The theory of affordances,'' \emph{Perceiving, acting and knowing: Towards an ecological psychology/Erlbaum}, 1977.

\bibitem{roboabc}
\BIBentryALTinterwordspacing
Y.~Ju, K.~Hu, G.~Zhang, G.~Zhang, M.~Jiang, and H.~Xu, ``Robo-abc: Affordance generalization beyond categories via semantic correspondence for robot manipulation,'' 2024. [Online]. Available: \url{https://arxiv.org/abs/2401.07487}
\BIBentrySTDinterwordspacing

\bibitem{industreal}
B.~Tang, M.~A. Lin, I.~Akinola, A.~Handa, G.~S. Sukhatme, F.~Ramos, D.~Fox, and Y.~Narang, ``Industreal: Transferring contact-rich assembly tasks from simulation to reality,'' \emph{arXiv preprint arXiv:2305.17110}, 2023.

\bibitem{admitlearn}
X.~Zhang, C.~Wang, L.~Sun, Z.~Wu, X.~Zhu, and M.~Tomizuka, ``Efficient sim-to-real transfer of contact-rich manipulation skills with online admittance residual learning,'' in \emph{Conference on Robot Learning}.\hskip 1em plus 0.5em minus 0.4em\relax PMLR, 2023, pp. 1621--1639.

\bibitem{atla}
A.~Z. Ren, B.~Govil, T.-Y. Yang, K.~R. Narasimhan, and A.~Majumdar, ``Leveraging language for accelerated learning of tool manipulation,'' in \emph{Conference on Robot Learning}.\hskip 1em plus 0.5em minus 0.4em\relax PMLR, 2023, pp. 1531--1541.

\bibitem{factory}
Y.~Narang, K.~Storey, I.~Akinola, M.~Macklin, P.~Reist, L.~Wawrzyniak, Y.~Guo, A.~Moravanszky, G.~State, M.~Lu \emph{et~al.}, ``Factory: Fast contact for robotic assembly,'' \emph{arXiv preprint arXiv:2205.03532}, 2022.

\bibitem{fmb}
J.~Luo, C.~Xu, F.~Liu, L.~Tan, Z.~Lin, J.~Wu, P.~Abbeel, and S.~Levine, ``Fmb: A functional manipulation benchmark for generalizable robotic learning,'' \emph{arXiv preprint arXiv:2401.08553}, 2024.

\bibitem{genchip}
K.~Burns, A.~Jain, K.~Go, F.~Xia, M.~Stark, S.~Schaal, and K.~Hausman, ``Genchip: Generating robot policy code for high-precision and contact-rich manipulation tasks,'' \emph{arXiv preprint arXiv:2404.06645}, 2024.

\bibitem{manipllm}
X.~Li, M.~Zhang, Y.~Geng, H.~Geng, Y.~Long, Y.~Shen, R.~Zhang, J.~Liu, and H.~Dong, ``Manipllm: Embodied multimodal large language model for object-centric robotic manipulation,'' in \emph{Proceedings of the IEEE/CVF Conference on Computer Vision and Pattern Recognition}, 2024, pp. 18\,061--18\,070.

\bibitem{imagemanip}
X.~Li, Y.~Wang, Y.~Shen, P.~Iaroslav, H.~Lu, Q.~Wang, B.~An, J.~Liu, and H.~Dong, ``Imagemanip: Image-based robotic manipulation with affordance-guided next view selection,'' \emph{arXiv preprint arXiv:2310.09069}, 2023.

\bibitem{variable}
Q.~Yang, A.~D{\"u}rr, E.~A. Topp, J.~A. Stork, and T.~Stoyanov, ``Variable impedance skill learning for contact-rich manipulation,'' \emph{IEEE robotics and automation letters}, vol.~7, no.~3, pp. 8391--8398, 2022.

\bibitem{impedance_irl}
X.~Zhang, L.~Sun, Z.~Kuang, and M.~Tomizuka, ``Learning variable impedance control via inverse reinforcement learning for force-related tasks,'' \emph{IEEE Robotics and Automation Letters}, vol.~6, no.~2, pp. 2225--2232, 2021.

\bibitem{compliancetuning}
X.~Zhang, M.~Tomizuka, and H.~Li, ``Bridging the sim-to-real gap with dynamic compliance tuning for industrial insertion,'' in \emph{2024 IEEE International Conference on Robotics and Automation (ICRA)}.\hskip 1em plus 0.5em minus 0.4em\relax IEEE, 2024, pp. 4356--4363.

\bibitem{deepwholebodycontrol}
Z.~Fu, X.~Cheng, and D.~Pathak, ``Deep whole-body control: learning a unified policy for manipulation and locomotion,'' in \emph{Conference on Robot Learning}.\hskip 1em plus 0.5em minus 0.4em\relax PMLR, 2023, pp. 138--149.

\bibitem{partnet}
K.~Mo, S.~Zhu, A.~X. Chang, L.~Yi, S.~Tripathi, L.~J. Guibas, and H.~Su, ``Partnet: A large-scale benchmark for fine-grained and hierarchical part-level 3d object understanding,'' in \emph{Proceedings of the IEEE/CVF conference on computer vision and pattern recognition}, 2019, pp. 909--918.

\bibitem{kirillov2023segment}
A.~Kirillov, E.~Mintun, N.~Ravi, H.~Mao, C.~Rolland, L.~Gustafson, T.~Xiao, S.~Whitehead, A.~C. Berg, W.-Y. Lo \emph{et~al.}, ``Segment anything,'' in \emph{Proceedings of the IEEE/CVF International Conference on Computer Vision}, 2023, pp. 4015--4026.

\bibitem{gsnet}
C.~Wang, H.~Fang, M.~Gou, H.~Fang, J.~Gao, C.~Lu, and S.~J. Tong, ``Graspness discovery in clutters for fast and accurate grasp detection,'' \emph{2021 IEEE/CVF International Conference on Computer Vision (ICCV)}, pp. 15\,944--15\,953, 2021.

\end{thebibliography}


\end{document}